\documentclass{article} 
\usepackage{paper,times}

\usepackage{hyperref}
\usepackage{url}

\title{Can an AI Win Ghana’s National Science and Maths Quiz? An AI Grand Challenge for Education}

\author{George Boateng\\
Kwame AI Inc., U.S.\\
ETH Zurich, Switzerland\\
\texttt{jojo@kwame.ai} \\
\And
Victor Kumbol \\
Kwame AI Inc., U.S. \\
Charite Berlin, Germany\\
\texttt{victor@kwame.ai} \\
\AND
Elsie Effah Kaufmann \\
University of Ghana, Ghana \\
\texttt{eeffahkaufmann@ug.edu.gh}
}

\iclrfinalcopy 
\begin{document}

\maketitle

\begin{abstract}
There is a lack of enough qualified teachers across Africa which hampers efforts to provide adequate learning support such as educational question answering (EQA) to students. An AI system that can enable students to ask questions via text or voice and get instant answers will make high-quality education accessible. Despite advances in the field of AI,  there exists no robust benchmark or challenge to enable building such an (EQA) AI within the African context. Ghana’s National Science and Maths Quiz competition (NSMQ) is the perfect competition to evaluate the potential of such an AI due to its wide coverage of scientific fields, variety of question types, highly competitive nature, and live, real-world format. The NSMQ is a Jeopardy-style annual live quiz competition in which 3 teams of 2 students compete by answering questions across biology, chemistry, physics, and math in 5 rounds over 5 progressive stages until a winning team is crowned for that year. In this position paper, we propose the NSMQ AI Grand Challenge, an AI Grand Challenge for Education using Ghana’s National Science and Maths Quiz competition (NSMQ) as a case study. Our proposed grand challenge is to “\textit{Build an AI to compete live in Ghana’s National Science and Maths Quiz (NSMQ) competition and win — performing better than the best contestants in all rounds and stages of the competition.}” We describe the competition, and key technical challenges to address along with ideas from recent advances in machine learning that could be leveraged to solve this challenge. This position paper is a first step towards conquering such a challenge and importantly, making advances in AI for education in the African context towards democratizing high-quality education across Africa.
\end{abstract}

\section{Introduction}
The COVID-19 pandemic highlighted and exacerbated the existing poor educational experiences of millions of students in Africa who were dealing with various challenges in their learning journey such as limited or no access to computers, the Internet, and qualified teachers.  In 2019, the average student-teacher ratio at the primary education level in Sub-Saharan Africa was 38:1 \cite{UNESCO2021} which is higher compared to 13.5:1  in Europe \cite{Eurostat2021}. Furthermore, only 65\% of the primary school teachers in Sub-Saharan Africa had the required minimum qualifications \cite{UNESCO2021}. This situation of lack of enough qualified teachers makes it challenging for students to have adequate learning support such as educational question answering. An AI system that could enable students in, for example, rural regions who only have access to feature phones and no access to the Internet to get instant answers to their questions via for example, toll-free calling would enable equitable and inclusive education, and dramatically democratize education across Africa.

Various AI teaching assistants (TA) have been developed for education question answering such as Jill Watson \cite{goel2016,goel2020}, Rexy \cite{benedetto2019}, a physics course TA \cite{zylich2020}, Curio SmartChat (for K-12 science) \cite{raamadhurai2019} and Kwame for Science (for Science in West Africa) \cite{boateng2022}. Yet, these AI systems have been evaluated in constrained settings (for example, text-only interactions) making them limited for real-world settings and hampering their potential to offer truly high-quality, scalable education. Hence, a  challenging benchmark that has the characteristics of the intended real-world use case is needed. 

Ghana’s National Science and Maths Quiz competition (NSMQ) is the perfect competition to evaluate the potential of such an AI. The NSMQ is a Jeopardy-style  annual live quiz competition in which 3 teams of 2 students compete by answering questions across biology, chemistry, physics, and math in 5 rounds over 5 progressive stages until a winning team is crowned for that year. Firstly, the NSMQ covers topics across chemistry, physics, biology, and mathematics, and features a wide range of question types such as calculation, multiple choice, true or false, riddles, etc., which encompass different types of questions students might ask an AI while learning. Secondly, the competition is highly competitive, with challenging questions that require critical thinking, and problem-solving skills making it ideal for evaluating an AI’s reasoning capabilities. Thirdly, expected answers to questions have different formats — short phrases, long phrases, numbers, etc., — which need different approaches of automatic evaluation in an AI system, some of which could be challenging such as for long answers. Finally, given the NSMQ is live, an AI would need to automatically transcribe spoken questions in real-world contexts, which will be critical for different learning contexts in which the AI could be deployed.

Consequently, we are proposing the \textbf{NSMQ AI Grand Challenge}, an AI Grand Challenge for Education using Ghana’s NSMQ competition as a case study. Our proposed grand challenge is to \textit{“Build an AI to compete live in Ghana’s National Science and Maths Quiz competition and win — performing better than the best contestants in all rounds and stages of the competition!”} Several AI grand challenges have been proposed over the years as a means to enable advances in AI \cite{reddy1988} with some already conquered. Some examples include Deep Blue winning Chess in 1997 \cite{deepblue}, Watson winning \textit{Jeopardy!} in 2011 \cite{watson}, and AlphaGo winning Go in 2016 \cite{alphago}. A recent grand challenge in education is the International Math Olympiad (IMO) Grand Challenge \cite{imo}. The NSMQ AI Grand Challenge differs from the IMO Grand Challenge by having a scope of both science and math education, consisting of a live competition (IMO is only written) with similarities to Jeopardy!, and importantly, centers on an African context.

The rest of this paper is as follows: in Section \ref{sec:overview} we describe the NSMQ competition, in Section \ref{sec:technical}, we describe key technical challenges to address for this competition, in Section \ref{sec:progress}, we describe the progress we have made and planned next steps, and we conclude in Section \ref{sec:conclusion}.

\section{Overview of NSMQ Competition}
\label{sec:overview}
The NSMQ is an annual live science and mathematics competition for senior secondary school students in Ghana. The quiz “\textit{aims to promote the study of the sciences and mathematics, help students develop quick thinking and a probing and scientific mind about the things around them while fostering healthy academic rivalry among senior high schools.}” \cite{nsmq}. The NSMQ was first run in 1993 and every year since then (1993 - 2022) except for 2010 and 2011 when the competition did not take place because of a lack of sponsorship. 

The format of the competition has evolved over the years. Currently,  in each contest, 3 teams of 2 students compete by answering questions across biology, chemistry, physics, and math in 5 rounds over 5 progressive stages until a winning team is crowned for that year. Teams have the chance to win cash and other prizes based on their performance in different rounds of the competition. The teams that participate in the NSMQ are chosen from schools across the country based on their performance in the regional competitions. In the regional competitions, schools compete to qualify for a specified number of slots, determined by the total number of eligible schools in the Region. Qualifying teams make it to the national competition. The national competition runs in 5 stages: the preliminaries, one-eighth, quarter-finals, semi-finals, and the grand finale. The school that wins the whole competition receives a trophy, a cash prize, and bragging rights until the next competition. The current quiz mistress of the competition is Professor Elsie Effah Kaufmann who is an Associate Professor of Biomedical Engineering and the Dean of the School of Engineering Sciences at the University of Ghana \cite{nsmq}. The questions for the competition are set by subject consultants who are academics from the University of Ghana. Though it is a high school competition, questions, especially beyond the quarterfinals, sometimes come from college-level content as well as knowledge of the history of science generally not within the standard Ghanaian high school syllabus.

\subsection{Round 1: Fundamental Concepts}
Round 1 is the round of Fundamental Concepts. In this round, teams are asked simple and direct questions that assess their understanding of fundamental concepts. A major question carries 3 points with partial points sometimes awarded. A wrongly answered question is carried over to the next 2 teams who can ring to answer for a bonus of 1 point and a penalty of -1 if they answer incorrectly.  Teams have 10 seconds to provide an answer and 30 seconds for questions that involve calculations. Some questions start with a preamble. Here is an example:

\textbf{Question 1:} 
A spectral lamp emits spectral lines with the color violet, blue-green, blue, and yellow-green. Rank the lines in increasing order of angular deviation through a prism with normal dispersion.

\textbf{Answer:}  Yellow-green, blue-green, blue, violet

\textbf{Question 2:}
A committee of 5 is to be formed from 7 students and 5 teachers. In how many ways can this be done if the committee has 3 students and 2 teachers?

\textbf{Answer:} 350 

\subsection{Round 2: Speed Race} 
Round 2 is the Speed Race. The questions are directed to all three teams simultaneously. For each question, each team gets a chance to answer by ringing their bell. Points are awarded as follows for correct answers: 3 points on the first attempt, 2 points on the second attempt, and 1 point on the third attempt. A wrongly answered question attracts a penalty of -1 point. Teams have 10 seconds to provide an answer and 30 seconds for questions that involve calculations. Some questions start with a preamble. Here is an example:

\textbf{Question 1:} 
The hanging of meat is a culinary process, commonly used in preparing beef for consumption. State one importance of this process. 

\textbf{Answer:} The process allows the enzymes to break down the muscle fibers in the meat which tenderizes the beef.

\textbf{Question 2:}
The name 2–ethylpentane is incorrect. What should be the systematic name for that compound?

\textbf{Answer:} 3–Methylhexane 

\subsection{Round 3: Problem of the Day}
Round 3 is known as the Problem of the Day. The contestants are required to solve a single question that usually has subparts within 4 minutes for a total of 10 points. They each write their answer on a public board for assessment by the quiz mistress. See Appendix \ref{sec:round3} for an example.

\subsection{Round 4: True or False}
Round 4 is known as the True or False round. Statements are presented to each team. The objective is for them to determine whether each statement is true or false. A correctly answered question fetches 2 points. A wrongly answered question attracts a penalty of -1 point. A team may decide not to answer a question, in which case the question will be carried over to the next contesting team that rings first for an opportunity to answer it as a bonus for the full benefit of the 2 points. Here is an example:

\textbf{Question:} 
Two obtuse angles cannot be supplementary.

\textbf{Answer:}  True

\subsection{Round 5: Riddles}
Round 5 is known as the Riddles round and arguably the most exciting round as the winner is generally determined by the performance in the round. In this round, 4 or more clues are presented to the teams that compete against each other to be first to provide an answer (which is usually a word or a phrase) by ringing their bell. The clues start vague and get more specific. To make it more exciting and encourage educated risk-taking, students get 5 points for getting it correct on the first clue, 4 points on the second clue, and 3 points for the third or subsequent clues. There are 4 riddles in all. Each riddle focuses on one of the 4 subjects. However, sometimes, the clues span different fields of science. See Appendix \ref{sec:round5} for an example.

\section{Technical Challenges}
\label{sec:technical}
In this section, we describe key technical challenges to be solved while leaning into ideas from recent advances in machine learning that could be leveraged to solve this grand challenge.

\subsection{Real-Time Speech Recognition}
The AI would need to accurately transcribe the questions being asked. Given the live format of the contest, it would need to distinguish the voice of the quiz mistress from those of contestants, work well for a Ghanaian accent, and correctly infer the start and end of the question even when questions have preambles. Furthermore, it would need to be robust to situations like the questions being repeated for the sake of clarity. Additionally, it would need to correctly transcribe scientific and mathematical terms/symbols in a format for textual analysis as well as terms peculiar to the Ghana setting (e.g., Ghana Cedis, etc.). It would also need to infer which is the current round of the competition to adequately format answers. Recently, Open AI released the Whisper model which was shown to work well for different accents and has shown some early promise \footnote{\url{https://github.com/openai/whisper}} \cite{radford2022}. Approaches would need to be implemented for speaker detection and diarization.

\subsection{Question Answering (QA)} 
The AI would need to formulate a textual answer to each question. One key challenge is that the answer needs to be framed in the expected format for each type of question in different rounds as follows: short answer with a word and phrase (fundamental concepts, speed race, riddles), true or false (true or false round), numbers and units (calculation type question in fundamental concepts, speed race, the problem of the day), and long answer (fundamental concepts, speed race, problem of the day). Potential approaches could entail retrieving relevant passages from a well-curated knowledge source using models such as Sentence-BERT \cite{reimers2020} and formulating the expected answer using models such as T5 \cite{raffel2019} for abstractive QA or models such as SciBERT \cite{beltagy2019} for extractive QA. Generative models which do not need a retrieval step such as  GPT-3 \cite{brown2020}, ChatGPT \cite{chatgpt} and Galactica \cite{taylor2022} could be explored though they are prone to producing factually inaccurate answers.
 
\subsection{Speech Output}
The AI would need to output a speech form of the textual answer obtained, using a local accent (e.g., Ghanaian). It would need to appropriately say answers for different scientific and mathematical textual representations. A recent model such as Microsoft’s VALL-E which can simulate a person’s voice could be used \cite{wang2023}.

\subsection{Data Collection}
Data from the competition would need to be curated and annotated for model development and evaluation. Recently, the competition has been live-streamed so the videos could be automatically transcribed and manually corrected. Past competitions that have been uploaded on YouTube could be retrieved. Furthermore, partnerships with the competition organizers could be set up to obtain the original version of the questions for all the years which may include hand-written documents. Considering the content is science and math, a lot of effort would be required to format it with the proper scientific and mathematical representation. 

\subsection{Model Evaluation}
Each predicted answer would need to be adequately evaluated against the correct answer for different answer types during model training and testing. True or false answers, short answers, and numeric answers are straightforward to evaluate via exact match. Long answers on the other hand are more challenging as current metrics like F1 (word overlap), ROGUE \cite{lin2003}, BLEU \cite{papineni2002}, BLUERT \cite{sellam2020}, METEOR \cite{banerjee2005}, BERTScore \cite{zhang2019} and Semantic Answer Similarity (SAS) \cite{risch2021} may not capture the nuances to properly assess the correctness of answers for a context like scientific question answering. Human-in-loop evaluation approaches have been proposed for science question answering \cite{bhakthavatsalam2021}. However, these are difficult and more time-consuming to implement. Better metrics would need to be developed.
 
\section{Progress and Plan}
\label{sec:progress}
We made some progress in the last few months of 2022 around the time the competition ran. Our system automatically transcribes a question, retrieves relevant passages from a knowledge source, and adapts the question for a particular round. First, we built Kwame AI \cite{boateng2022} \footnote{\url{https://kwame.ai/}}, an AI platform that enables students to ask questions related to the West African Science curriculum and get instant answers from a well-curated knowledge source consisting of textbooks and relevant past national exam questions related to their query. It also enables developers to make API calls with questions and receive relevant passages as answers. Second, we implemented a basic real-time speech-to-text system that can transcribe spoken questions using Open AI’s Whisper model \footnote{\url{https://youtu.be/Eozpv0IkJBo}}. We then integrated it with Kwame AI by making API calls with the transcribed question allowing Kwame to retrieve relevant passages in response to questions \footnote{\url{https://youtu.be/32E6bl6WzjQ}}. Finally, we explored adaptation for one of the rounds — Round 5: Riddles as it is easier to automatically evaluate whether the AI is right or wrong (since the answers are usually just one word or a short phrase), and it is also the most exciting round. We used the SciBERT model, which was pretrained on scientific text to extract a span of text from passages returned in response to our API queries of riddle clues \footnote{\url{https://youtube.com/shorts/skKqKFPeqA4?feature=share}}. As these are exploratory progress, robust and thorough evaluations are needed, as well as further processing of these passages to arrive at different kinds of answers for different rounds (calculation types, true or false, riddles, etc.).

Our current concrete minimal goal is to have the first NSMQ AI deployed and working for at least one of the 5 rounds of the competition — Round 5: Riddles in time for the 2023 NSMQ which will take place around Fall 2023. First, we are inviting individuals to contribute to this challenge via an open-source project in both technical (software engineering, ML, etc.) and non-technical (e.g., marketing, data curation, etc.) ways. Second, we plan to launch an official Kaggle-like competition with our baseline system and results to crowdsource contributions for improvement. We will seek partnerships and sponsorship from organizations and companies in various ways such as to fund the prize money for the competition, provide cloud compute resources, etc.

\section{Conclusion}
\label{sec:conclusion}
In this position paper, we proposed the NSMQ AI Grand Challenge to enable the advancement of real-world AI systems for education in the African context.  Unfortunately, the African continent tends to be an afterthought in terms of applications of AI advances. This work seeks to center the building and advancement of AI for Education that works well first and foremost in the African context. The NSMQ competition is an exciting one with interesting technical challenges whose solutions have real-world impact potential in education. An AI capable of such a feat can offer individual learning assistance at scale to millions of students and enable accessible, equitable, high-quality education.

\clearpage
\bibliography{paper}
\bibliographystyle{paper}

\appendix
\clearpage

\section{Round 3: Problem of the Day Example}
\label{sec:round3}
\textbf{Question:} 
To determine the sensitivity of the body to temperature, a Senior High School student performed the experiment below: 

He collected three jars or beakers of about the same size and filled one with cold water (10-15 °C), one with hot water (40-50 °C), and the third with warm water (about 25 °C).  The first finger of the left hand was placed in the cold water and the first finger of the right hand in the hot water.  He left both fingers immersed for at least one minute. After one minute, he removed both fingers from the jars and dipped them repeatedly but alternately in the warm water for about a second at a time.

What impression did (i) the left finger, and (ii) the right finger give about the temperature of the warm water? 

Why should there be any difference in the sensory information from the two fingers?

How could you modify the experiment to test your suggestion in ‘2’ above? 

Does the result mean that the skin of the fingers is not capable of judging whether an object is hot or cold? 

What does the result suggest about the way in which the skin responds to temperature?

\textbf{Answer:}  
i. The finger (left) which has been immersed in cold water will register warmth. 
ii. The finger previously held in hot water will register coldness. 

The difference in sensations can be attributed to the difference in the temperature of the fingers after one minute's immersion. / An alternative explanation is that the warmth receptors in the hot water become adapted, i.e. after prolonged immersion they no longer send impulses to the brain. Consequently, on transfer to lukewarm water, there are few impulses sent to the brain from the warmth receptors of this finger, whereas the warmth receptors in the cold finger fire normally.

This can be tested by repeating the experiment with the left-hand finger in the hot water and the right-hand finger in the cold water. The sensation upon dipping them both into warm water should be the reverse of the first experiment.
  
The results suggest that the fingers detect whether they are gaining or losing heat rather than the actual temperature of an object. Metal objects at room temperature will feel cold to the touch because heat is conducted away from the fingers, while wooden objects at the same temperature feel less cold. 

The results seem to imply that the thermo-receptors respond to changes in temperature rather than to any particular temperature. In fact, there is a steady discharge of nerve impulses from cold and warmth receptors at all temperatures within certain limits but increased bursts of impulses occur during sudden changes in temperature.

\section{Round 5: Riddles Example}
\label{sec:round5}
\textbf{Question:} 
\begin{enumerate}
\item I am a property of a periodic propagating disturbance.
\item Therefore, I am a property of a wave.
\item I describe a relationship that can exist between particle displacement and wave propagation direction in a mechanical wave. 
\item I am only applicable to waves for which displacement is perpendicular to the direction of wave propagation.
\item I am that property of an electromagnetic wave which is demonstrated using a polaroid film.
\end{enumerate}

Who am I?

\textbf{Answer:} Polarization

\end{document}